\documentclass{article}

\PassOptionsToPackage{numbers, compress}{natbib}

\usepackage[preprint]{neurips_2026}

\usepackage[utf8]{inputenc}
\usepackage[T1]{fontenc}
\usepackage{hyperref}
\usepackage{url}
\usepackage{booktabs}
\usepackage{amsfonts}
\usepackage{nicefrac}
\usepackage{microtype}
\usepackage{xcolor}
\usepackage{amsmath}
\usepackage{amssymb}
\usepackage{graphicx}
\usepackage{multirow}
\usepackage{makecell}
\usepackage{caption}
\usepackage{subcaption}
\usepackage{colortbl}
\usepackage{algorithm}
\usepackage{algpseudocode}
\usepackage{wrapfig}
\usepackage{float}
\usepackage{xcolor}

\newcommand{\secref}[1]{Section~\ref{#1}}
\newcommand{\figref}[1]{Figure~\ref{#1}}
\newcommand{\tabref}[1]{Table~\ref{#1}}
\newcommand{\appref}[1]{Appendix~\ref{#1}}

\graphicspath{{figures/}}


\definecolor{expdisgreen}{RGB}{31, 120, 65}
\definecolor{expdisdark}{RGB}{20, 80, 45}
\definecolor{headlineyellow}{RGB}{255, 245, 200}

\definecolor{softgray}{RGB}{240, 240, 240}
\definecolor{deltagray}{RGB}{115, 115, 115}
\definecolor{linkblue}{RGB}{0, 45, 140}
\hypersetup{colorlinks=true, linkcolor=linkblue, citecolor=linkblue, urlcolor=linkblue}

\hypersetup{
    colorlinks=true,
    linkcolor=blue,
    filecolor=magenta,      
    urlcolor=cyan,
    citecolor=blue,
    pdftitle={Decoupling Exploration from Optimization in RLVR},
    pdfpagemode=FullScreen,
    }

\title{Decoupling Exploration from Optimization in RLVR}

\author{%
  Saif Punjwani$^{1,\ast}$ \;
  Micah Goldblum$^{1,2,\ast}$ \\[4pt]
  Columbia University \\
}

\begin{document}

\maketitle
\renewcommand{\thefootnote}{\fnsymbol{footnote}}%
\footnotetext[1]{Corresponding authors: \texttt{saif.punjwani@columbia.edu}, \texttt{micah.g@columbia.edu}.}%
\renewcommand{\thefootnote}{\arabic{footnote}}%
\footnotetext[1]{Department of Computer Science, Columbia University.}%
\footnotetext[2]{Department of Electrical Engineering, Columbia University.}%
\footnotetext[3]{We release all \href{https://github.com/SaifPunjwani/Exploration-Distillation}{code} and \href{https://huggingface.co/SaifPunjwani/expdis-checkpoints}{checkpoints}.}

\begin{abstract}
Modern language models undergo reinforcement learning with verifiable rewards (RLVR) on top of already-trained checkpoints. A key promise of RLVR is the discovery of new reasoning strategies. In principle, a model can sample novel ideas absent from its prior training data. In practice, however, augmenting RLVR with strong novelty incentives has seen limited success and can degrade model quality. Because verifiable rewards supervise only a narrow slice of the model's knowledge and behavior, such degradations are difficult to recover from. Instead, we decouple exploration from optimization in a framework we call Exploration-Distillation (ExpDis). We train one or more \textit{explorer} policies with a novelty bonus in the reward, filter their trajectories for correctness and quality, and distill them into a separate \textit{student} policy. The student policy is then trained without a novelty bonus. We repeat the above procedure for several rounds, alternating between exploration and optimization. This decoupling allows us to aggressively scale exploration without degrading the student policy.
Across seven mathematical reasoning benchmarks and two model families, ExpDis outperforms DAPO at the same wall-clock budget. Moreover, we observe improved pass@$k$ scaling, indicating that ExpDis produces models that generate more diverse correct solutions.\footnotemark[3]
\end{abstract}

\section{Introduction}
\label{sec:intro}

Reinforcement learning with verifiable rewards (RLVR) has become the dominant recipe for improving the reasoning abilities of language models~\citep{lambert2024tulu3,guo2025r1,shao2024deepseekmath,liu2025drgrpo,yu2025dapo}. RLVR samples solutions to problems, scores each solution with a programmatic verifier, and shifts probability toward those that pass. By iteratively sampling new solutions and reinforcing successful ones, RLVR can in principle discover and then reinforce new solutions that were absent from the model’s training data.  In practice, however, it can only reinforce what the model samples, so discovery depends on exploration.

A natural way to boost exploration is to add an explicit novelty bonus to the reward, following a long line of intrinsic-motivation methods in RL~\citep{pathak2017curiosity,bellemare2016unifying,burda2018rnd,eysenbach2019diayn}. When the model explores new solution strategies, some of them may succeed but often most of them fail, leading to a temporarily degraded model as we now are left with a model that mostly generates failed solutions.
Though our initial policy is a pretrained LLM with vast knowledge including virtually all of Wikipedia, the reward supervises only a tiny slice of its knowledge. This means an LLM could forget most of Wikipedia and still achieve high reward. The reward therefore cannot detect damage that exploration causes outside that slice, let alone repair such degradations. We argue that this failure is not a property of novelty as a signal, but of where the novelty bonus is applied.

\begin{figure}[t]
  \centering
  \includegraphics[width=\linewidth]{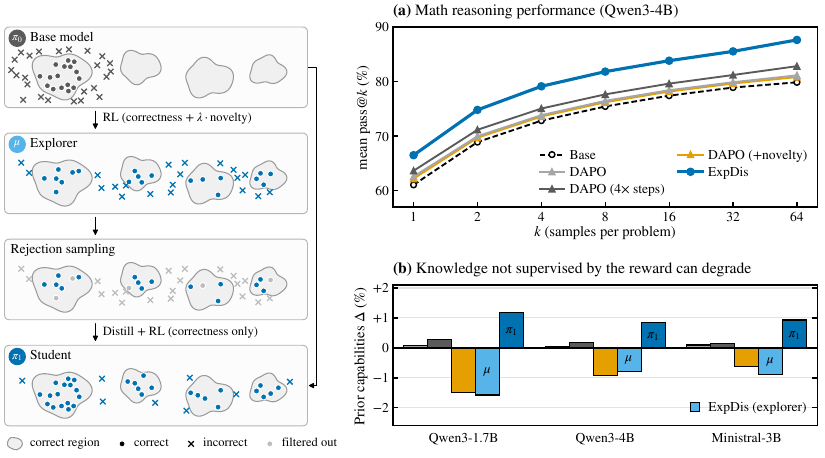}
  \caption{\textbf{Decoupling exploration from optimization improves math accuracy without degrading prior capabilities.} \textbf{Left:} Rather than coupling exploration and optimization in a single RLVR run, ExpDis assigns these roles to separate policies. We first train an explorer $\mu$ with a novelty bonus in its reward, then distill a diverse pool of its reasoning traces to cold start a student, and finally train the student with standard RLVR using only a correctness reward. As depicted in the figure, this allows us to explore a diverse set of reasoning strategies and preserve this diversity in our student policy. \textbf{(a)} ExpDis outperforms DAPO in both average accuracy and pass@$k$, even when DAPO is trained for $4\times$ as long (Qwen3-4B, mean pass@$k$ over five math benchmarks). \textbf{(b)} Adding a novelty bonus (RND) directly to the DAPO reward degrades performance on prior capabilities relative to the base model. Because the reward supervises only a narrow slice of the model's knowledge (here, mathematical reasoning), achieving high reward does not prevent general capability degradation. By decoupling the explorer from the student policy, we avoid this degradation.}
  \label{fig:headline}
\end{figure}

We decouple exploration and optimization by giving them separate model instances, both trained from the same base checkpoint. An \textit{explorer policy} explores aggressively under a novelty bonus whose weight $\lambda$ sets how strongly it is pushed toward unfamiliar solutions relative to correctness. A separate \textit{student policy} never sees a novelty reward. It learns from the explorer only through filtered trajectories, and is then trained with standard RLVR. The two models communicate through filtered trajectories rather than parameters, so the student policy absorbs the explorer's discoveries without inheriting its degradations induced by exploration. We call this framework \textit{Exploration-Distillation}.

Decoupling makes exploration safe to scale aggressively, since the student policy never receives the novelty bonus and the filter keeps failed trajectories out of its training data. We can scale \emph{breadth} by running several explorers in parallel and distilling their pooled trajectories into one student policy. We can also scale \emph{depth} by repeating the pipeline for multiple rounds that each start from the previous student. We scale both of these axes independently and find that their composition enables sustained performance gains even when all experiments are wall-clock matched.

We evaluate ExpDis on mathematical reasoning benchmarks with Qwen3-1.7B, Qwen3-4B, and Ministral-3-3B-Instruct-2512 as base models. At the same amount of compute, ExpDis outperforms DAPO in both pass@1 and pass@64. Furthermore, spreading this fixed budget over multiple parallel explorers and several rounds of the ExpDis procedure yields further gains. Adding a novelty bonus directly to DAPO, in contrast, increases diversity but degrades performance on general knowledge benchmarks. In ExpDis, the explorer can be pushed to far higher diversity without harming the student, which improves in both diversity and accuracy and avoids degradation.

In principle, ExpDis is compatible with any exploration mechanism, since the student learns only from the explorer's filtered trajectories. Since the explorer is never deployed, it can also use mechanisms too aggressive to apply to the deployed model directly. Exploration methods that would destabilize standard RLVR may therefore become viable under ExpDis.

\section{Related Work}
\label{sec:related}

\paragraph{Exploration in LLM reasoning.} A number of works ask whether standard RLVR algorithms discover new solutions or only sharpen a model's pre-RL generative distribution. \citet{shao2024deepseekmath} observe that RL improves majority-vote accuracy but not pass@$k$, and \citet{chen2026does} find that base models overtake their RLVR-trained counterparts at large $k$, suggesting that RLVR narrows the set of problems a model can solve. Others find that RL can expand this set through prolonged training~\citep{liu2025prorl}, or that the model learns new skills by composing existing ones~\citep{yuan2025compose}. A related line of work studies entropy collapse: policy entropy falls sharply early in training~\citep{cui2025entropy}, and the gains from RLVR rely on a small set of high-entropy tokens to allow sampling diverse solutions~\citep{wang2026beyond}. Many methods counteract this entropy collapse during training. For example, DAPO~\citep{yu2025dapo} raises the upper clipping bound so that low-probability tokens with positive advantage can grow faster, whereas \citet{skywork-or1-2025} directly regulate the policy's entropy. Other approaches add exploration or novelty bonuses to the reward~\citep{outcome2024outcome,imentor2025,tuyls2025representation,li2025darling}. These methods draw on a long line of exploration bonuses in classical RL, including count-based bonuses~\citep{bellemare2016unifying}, intrinsic curiosity~\citep{pathak2017curiosity}, and random network distillation~\citep{burda2018rnd}.  ExpDis is compatible with a broad spectrum of exploration bonuses, and we try plugging several example bonuses into our framework, observing benefits of exploration in all cases.

\paragraph{Reasoning trace distillation.} A classical precursor that trains a policy to imitate a separate process that finds better solutions is expert iteration~\citep{anthony2017exit}, in which a tree-search expert generates improved moves and a neural apprentice learns to imitate them. Today, distilling reasoning traces from a stronger model is standard practice: DeepSeek-R1~\citep{guo2025r1} is distilled into smaller and faster open models, and several works study how to curate reasoning traces to instill reasoning in fine-tuned models~\citep{ye2025limo, guha2026openthoughts}. When no stronger model is available, a model can instead learn from its own filtered samples, as in STaR~\citep{zelikman2022star}, ReST~\citep{gulcehre2023rest}, and ReST$^{\text{EM}}$~\citep{singh2023beyond}. DeepSeek-R1 fine-tunes its base model on correct traces rejection-sampled from an RL checkpoint before running RL again~\citep{guo2025r1}, and multi-teacher on-policy distillation methods~\citep{ma2026mopd} distill separately trained specialist models into one model. We similarly distill trajectories from one model into another, but we train our explorer policies from the same base model as our student with the distinct goal of exploration.

\paragraph{Decoupling exploration in RL.}
Separating exploration from exploitation has a long history in RL: E$^3$~\citep{kearns2002near} keeps separate exploration and exploitation policies and explicitly switches between them, \citet{csimcsek2006intrinsic} explore with a policy rewarded for improving a separate task value function that never sees this reward, and Go-Explore~\citep{ecoffet2021goexplore} first finds high-reward trajectories and then turns them into a robust policy via imitation learning.
In reinforcement learning for LLMs, R$^2$PO~\citep{r2po2025} trains a residual logit head to diversify within-group outcomes, uses it to sample rollouts for the policy's GRPO updates, and discards it at inference. In concurrent work, ICE~\citep{dilkes2026ice} explores with instruction-conditioned prompts and self-distills correct rollouts into the unconditioned policy. In both, the exploring policy and the student share weights. We instead train the explorer as a separate network and pass only filtered trajectories to the student.  This design choice insulates the student from exploration pressures and thus avoids degrading the student.

\section{Method}
\label{sec:method}
Exploration-Distillation trains an explorer policy $\mu$ with a novelty bonus to generate diverse trajectories, filters them for quality and correctness, and distills the resulting trajectories into a student policy $\pi_1$. We present a single iteration below and describe how we scale it in \secref{sec:multiscout} and \secref{sec:multiround-method}. We study the importance of each stage in \appref{app:ablations}.

\paragraph{Explorer policy training.}
\label{sec:scout}
We initialize an explorer policy $\mu$ from the base policy $\pi_0$ with DAPO \citep{yu2025dapo} and augment its reward with a novelty reward $r_\text{novelty} \ge 0$.
\begin{equation}
\label{eq:scout-reward}
  r_\text{explorer}(\tau) \;=\; r_{\text{correct}}(\tau)
  \;+\; \lambda \cdot r_{\text{novelty}}(\tau) \cdot \mathbf{1}\!\left[r_{\text{correct}}(\tau) = +1\right]
  \;+\; r_{\text{overlong}}(\tau),
\end{equation}
where $r_{\text{correct}} \in \{-1, +1\}$ and $\lambda \geq 0$. As introduced in DAPO, $r_{\text{overlong}} \in [-1, 0]$ is a soft penalty on completions that approach the length budget.  Apart from the novelty term, the explorer is identical to our DAPO baseline.  This reward formulation ensures that every correct rollout receives a higher reward than any incorrect one for any value of $\lambda$. We find that applying the bonus to all rollouts leads to training instability as a model can achieve increased reward simply by generating novel garbage.

\paragraph{Rejection sampling.}
\label{sec:filter}
We now need a method for collecting trajectories generated by the explorer policy to be distilled into the student.  We remove any incorrect trajectories, since the point of this rejection sampling step is to insulate the student from any degradations caused to the explorer by novelty pressure.  A naive choice would be to take all correct explorer trajectories or a random subset and hand them off, but we find that filtering not only for correctness but also for quality strongly improves performance. A trajectory passes the filter only if (1) its final \texttt{\textbackslash boxed\{\}} answer is correct, (2) it terminates within a fixed token budget, and (3) it contains no repetition loops.

After this filtering, we keep only a single shortest resulting trajectory per problem. Problems that the explorer solves correctly many times therefore do not dominate the student's training data. We cap the resulting set at 500 trajectories per round. In \appref{app:ablations}, we ablate our quality filtering procedure.

\paragraph{Student policy training.}
\label{sec:central}
We distill the base policy on the resulting trajectories simply by performing supervised finetuning (SFT) on those trajectories with the standard cross-entropy loss. Following SFT, we train the distilled model with DAPO and the standard correctness-only reward (i.e., no novelty bonus) to obtain the student policy $\pi_1$, which inherits none of the degradation from exploration pressure.  The student both receives the benefit of exploration via distillation, which is off-policy, and also the benefit of on-policy training during its DAPO stage.

\begin{figure}[t]
\centering
\includegraphics[width=\linewidth]{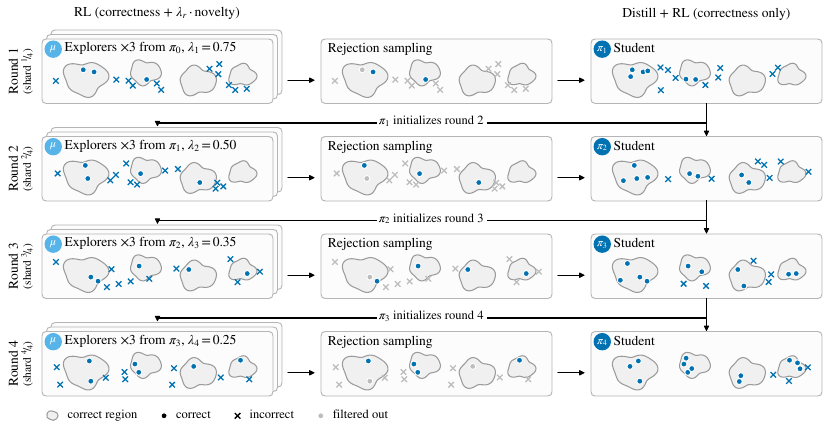}
\caption{\textbf{ExpDis scales across parallel explorers and iterations.} Within a round, several explorers are trained in parallel with a novelty bonus, and their traces are pooled to distill a single student. Across rounds, that student initializes both the next round's explorers and the next round's student, so every round starts from a stronger model. Each round trains on its own shard of the training prompts, and the novelty weight anneals from $\lambda_1{=}0.75$ to $\lambda_4{=}0.25$. We study how to allocate a fixed RL budget across explorers and rounds in \secref{sec:results}.}
\label{fig:mrme-diagram}
\end{figure}

\subsection{Parallel Explorer Policies}
\label{sec:multiscout}
Rather than train a single explorer policy $\mu$, we can train $K$ independent explorer policies and pool their trajectories before filtering and performing distillation. Compared to the single-explorer variant, training parallel explorers can increase the diversity of trajectories we curate for the student. This diversity increases coverage over solutions and boosts our probability of discovering successful ones.  

We set the novelty bonus weight $\lambda$ identically for all peer explorer policies, as we find little benefit from varying them. In our experiments, we divide compute among explorers, requiring no additional wall-clock time over a single explorer.  The more compute a training run uses, the more parallel explorers we expect to be optimal.

\subsection{Iterative Exploration-Distillation}
\label{sec:multiround-method}

We can scale the above method by performing many rounds of alternating exploration and distillation.  After producing a student policy, we can use this student as a strong initialization for both the next explorer policy and subsequent student policy. Formally: to train student $\pi_t$, we first train one or more explorers $\mu$ from $\pi_{t-1}$, filter the trajectories, distill those trajectories onto $\pi_{t-1}$ via SFT and then perform standard RLVR on the distilled checkpoint without an exploration bonus to produce $\pi_t$.

We partition the training prompts into $R$ non-overlapping shards $\mathcal D_1, \dots, \mathcal D_R$ of equal size. As shown in \figref{fig:mrme-diagram}, round $r$ draws its training prompts from $\mathcal D_r$. Hence, each round pairs a stronger initialization with prompts it has not yet trained on.

\paragraph{Annealing novelty bonus.} Because exploration is split across several rounds, we can set its strength per round through the novelty-bonus weight $\lambda$, which we anneal on a fixed schedule from $0.75$ to $0.25$ over four rounds. This schedule follows the common practice of decreasing exploration pressure as training progresses~\citep{sutton1998reinforcement}. Early rounds, where correct trajectories are still undiscovered, get strong novelty pressure whereas later rounds get lower novelty pressure to consolidate what has been found.

\section{Experimental Setup}
\label{sec:setup}

\paragraph{Base models.}
\label{sec:setup-models}
We evaluate ExpDis on Qwen3-1.7B, Qwen3-4B~\citep{qwen3}, and Ministral-3-3B-Instruct-2512~\citep{ministral3}. The Qwen3 models are distilled from larger RL-trained Qwen3 models rather than trained with RL directly~\citep{qwen3}, whereas Ministral-3-3B-Instruct is instruction-tuned and undergoes online preference optimization~\citep{ministral3}. We describe all sampling parameters in \appref{app:contract}.

\paragraph{Novelty bonus.} In principle, ExpDis is compatible with any novelty bonus.  In our primary experiments, we choose a Random Network Distillation (RND) bonus \citep{burda2018rnd} as we find that this formulation is especially effective.  Following \citet{burda2018rnd}, we train one small predictor MLP $f_{\theta,\ell}$ per layer to match a frozen random target MLP $f^{\star}_{\ell}$ on features extracted from the language model's own hidden states, and use the prediction error as the novelty signal. We extract features from a fixed set of intermediate layers $\mathcal L$, mean-pool over completion tokens (the completion is encoded without the prompt), and average across layers:
\begin{equation}
\label{eq:rnd}
  r_{\text{novelty}}(\tau) \;=\; \frac{1}{|\mathcal L|}\!\sum_{\ell \in \mathcal L}
   \frac{1}{\sqrt{d}}\,\bigl\lVert f^{\star}_{\ell}\!\bigl(h_{\ell}(\tau)\bigr) - f_{\theta,\ell}\!\bigl(h_{\ell}(\tau)\bigr) \bigr\rVert_2.
\end{equation}
Here $h_{\ell}(\tau)$ is the pooled layer-$\ell$ feature of trajectory $\tau$ and $d$ is the output dimension of the predictor and target networks, so each layer contributes a root-mean-square prediction error. The predictor updates continuously on the same features that drive the bonus, while the target stays frozen for the entire run. In preliminary experiments, features from several intermediate layers gave a cleaner signal than last-layer features alone (layers and network sizes in \appref{app:contract}). We ablate the choice of $r_\text{novelty}$ in \appref{app:novelty-swap} and observe a consistent improvement over DAPO across bonuses.

\paragraph{Reward.} We use a standard math verifier for $r_\text{correct}$. In \appref{app:novelty-bonus}, we sweep the weight $\lambda$ on $r_\text{novelty}$ and find that even very large weights leave the student model outperforming the base model. We find that performance roughly peaks at $\lambda$=0.5 and conduct all experiments with this value of $\lambda$.

\paragraph{Training dataset.}
\label{sec:setup-data}
We sample random prompts without replacement from DAPO-Math-17K~\citep{yu2025dapo}, a collection of 17K mathematical reasoning problems, until reaching a cap of 300 steps (discussed below). We perform additional experiments on the DeepScaleR~\citep{luo2025deepscaler} dataset in \appref{app:ablations}.

\paragraph{Evaluation.}
\label{sec:setup-benchmarks}

We report performance on seven mathematical reasoning benchmarks: AIME24, AIME25, AIME26~\citep{maa2025amc}, MATH500~\citep{hendrycks2021math,lightman2023verify}, AMC23~\citep{maa2025amc}, Minerva-Math~\citep{lewkowycz2022minerva}, and GSM8K~\citep{cobbe2021gsm8k}. Following the Qwen3 technical report~\citep{qwen3}, sample $64$ completions per problem (32 for AMC23 and 8 for GSM8K) and report the unbiased pass@$k$ estimator for $k\in\{1,2,4,\ldots,64\}$. GSM8K is close to saturation for all three models, and AMC23 is close to saturation for the Qwen3 models, which leaves little room to compare methods. We therefore focus on the other five benchmarks and report their mean unless noted otherwise.

We additionally study the model's degradation on a suite of general capability benchmarks: MMLU-Pro~\citep{wang2024mmlu}, MMLU-Redux~\citep{gema2025we}, and GPQA-Diamond~\citep{rein2023gpqa} for knowledge, ZebraLogic~\citep{lin2025zebralogic} for logical reasoning, and IFEval~\citep{zhou2023instruction} for instruction following. This serves as a proxy for capabilities not directly supervised by our training dataset and reward to study the extent to which exploration pressures can degrade a model's knowledge over time.

\paragraph{Baselines.}
\label{sec:setup-baselines}
We compare against three RLVR baselines with a standard correctness-only reward: GRPO~\citep{shao2024deepseekmath}, Dr.~GRPO~\citep{liu2025drgrpo}, and DAPO~\citep{yu2025dapo}. We find empirically that DAPO is the strongest of these baselines and design two extensions: (1)~adding a novelty bonus to the reward with $\lambda{=}0.5$ to isolate the effect of decoupling exploration, and (2)~running DAPO for $4\times$ as many steps.

\paragraph{Compute matching.}
We perform 300 steps of RL with 4 prompts per step, each having a group size of 16 and a 32,768 max generation limit. We also study the effect of training DAPO for $4\times$ the number of steps to determine if prolonged training allows baselines to catch up \citep{liu2025prorl}.

ExpDis splits these updates between the two policies: 200 for the explorer and 100 for the student. When we scale across explorers or rounds, each share is divided evenly among them, while we keep the SFT stage of the student policy two epochs on at most 500 filtered explorer trajectories per round.

\begin{figure}[t]
\centering
\includegraphics[width=\linewidth]{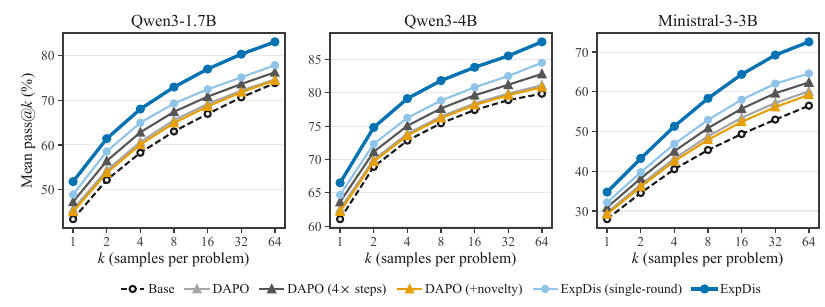}
\caption{\textbf{ExpDis outperforms all baselines under the same RL compute budget.} By partitioning this budget into alternating rounds of exploration and optimization (as described in \secref{sec:multiround-method}), we can surpass DAPO and even a single round of ExpDis. We report full benchmark results in \appref{app:extended-results}.}
\label{fig:passk-mean}
\end{figure}

\section{Results}
\label{sec:results}

Across all three models, ExpDis outperforms the DAPO baseline in mean accuracy on mathematical reasoning benchmarks (\figref{fig:passk-mean}). A single round of ExpDis already outperforms DAPO trained for $4\times$ as many steps. By spreading the same compute over several rounds of alternating exploration and optimization (\secref{sec:multiround-method}), we observe further gains. All ExpDis runs outperform this prolonged DAPO run in both average performance and pass@$k$ up to 64 generations per problem.

\begin{figure}[b]
    \centering
    \includegraphics[width=1\linewidth]{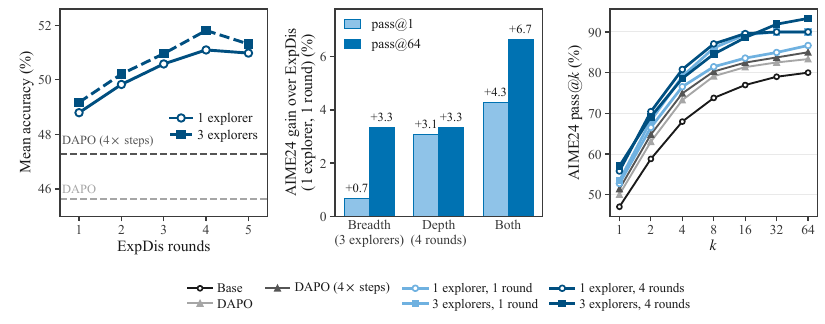}
    \caption{\textbf{Under a fixed RL compute budget, how we allocate compute impacts downstream student policy performance.} We vary the number of explorers and rounds under a fixed compute budget, and find complementary impacts of each scaling axis. Under a fixed budget, adding rounds improves pass@1 more than adding explorers, and combining both gives the largest gains in pass@1 and pass@64.  All results are reported on Qwen3-1.7B.}
    \label{fig:scaling}
\end{figure}

\paragraph{Scaling exploration.} Decoupling lets us scale exploration along two axes: running several explorers in parallel within a round and repeating the ExpDis procedure over several rounds. In \figref{fig:scaling}, we vary both on Qwen3-1.7B while holding the total budget fixed at 300 RL updates. We observe that depth contributes more to pass@$1$ than breadth, though a combination of both improves pass@$k$ over either. One interpretation is that the two axes play different roles: each additional round reinforces a progressively stronger student, whereas additional explorers widen the set of problems with correct trajectories and the variety of trajectories that get distilled into the student.

At our budget, gains saturate and begin to show signs of regression after three explorers and four rounds. We expect this optimum to shift as the budget grows: with a higher compute budget, longer rounds and larger batch sizes per explorer will have diminishing marginal returns, and we expect the optimal number of rounds and explorers to increase accordingly. Note that we vary breadth and depth at a fixed compute budget; explorers run in parallel, and spreading compute across more explorers adds no wall-clock time.  In principle, this parallelization allows practitioners with vast compute access to ramp up the number of explorers at no time cost.

\paragraph{Diversity analysis.} RLVR is known to collapse entropy and sharpen the sampling distribution as training proceeds~\citep{cui2025entropy,chen2026does}. Counteracting this collapse within a single model requires a delicate balance, since the same model must both explore diverse reasoning strategies but must simultaneously be optimized for correctness and avoiding incorrect solutions. Adding a novelty bonus to DAPO illustrates this trade-off: it increases both lexical and answer diversity relative to standard DAPO, yet slightly reduces mathematical reasoning performance.

\begin{figure}[h]
    \centering
    \includegraphics[width=1\linewidth]{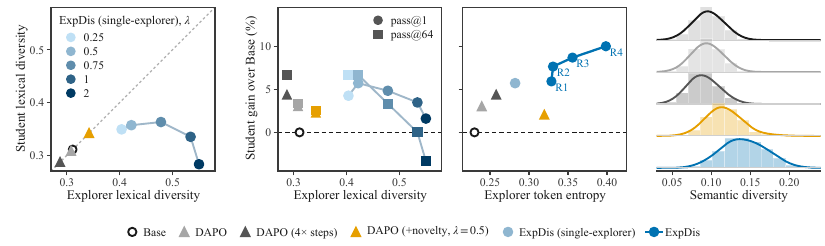}
    \caption{\textbf{Explorer policy can be driven to more diversity without harming the student.} \textbf{Left:} We study how explorer policy diversity impacts the downstream student in ExpDis. In general, we find that we can push the diversity of the explorer policy without corrupting the downstream student policy. Since we filter for correctness and quality before distilling, a moderate increase in explorer diversity (for example, lexical diversity) brought on by tuning the novelty bonus weight $\lambda$ does not negatively impact the student policy. Similarly, we can increase the explorer token entropy far beyond that of DAPO (+novelty) without incurring downstream reasoning degradation. \textbf{Right:} We find that ExpDis generations contain more semantically diverse mathematical reasoning strategies. We define each diversity metric in \appref{app:diversity}.}
    \label{fig:diversity} 
\end{figure}

Diversity and performance are not always complementary; a model may, for example, learn to rephrase the same solution strategy, thereby increasing lexical diversity or entropy without improving pass@$k$. As shown in \figref{fig:diversity}, DAPO + novelty exhibits higher lexical diversity and token entropy, but regresses on mathematical reasoning compared to the DAPO model. Broadly, we find that we can push the explorer toward far greater diversity than a single model could tolerate while maintaining or improving on downstream accuracy.

\paragraph{Stage ablations.} We run ablations on Qwen3-1.7B to understand whether ExpDis reduces to either (1) distilling unfiltered explorer trajectories and then running RL, or (2) distilling filtered trajectories without running RL afterward. We find that both variants score slightly below DAPO on AIME24, whereas the full procedure scores above it (\appref{app:ablations}). Furthermore, keeping only the shortest valid trajectory per problem (i.e., quality filtering) outperforms keeping every correct trajectory. Finally, we replace RND with a kNN-based or an elliptical novelty bonus (the latter adapted from \citet{tuyls2025representation}) and find that mean accuracy drops by at most $1.6$ points, and both variants still outperform DAPO (\appref{app:novelty-swap}).

\begin{wrapfigure}{r}{0.35\linewidth}
\vspace{-0.5em}
\centering
\includegraphics[width=\linewidth]{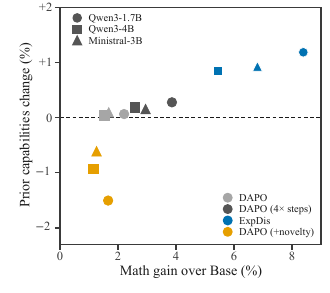}
\caption{Adding RND novelty bonus to DAPO degrades prior capabilities.}
\label{fig:novelty-tax}
\vspace{-3em}
\end{wrapfigure}
\paragraph{Novelty bonus degradation.} A verifiable reward, typically applied to math and code tasks, directly supervises only a narrow slice of a model's capabilities. In our experiments, DAPO with the RND novelty bonus degrades below the base model on prior capability benchmarks (\secref{sec:setup}) despite improving on math (right). Conversely, ExpDis gets the benefits of exploration without degradation. We report results for each general capability benchmark in \figref{fig:novelty-tax-breakdown}.

While the drops we observe are modest at our scale, we speculate that larger-scale RLVR runs that apply exploration bonuses directly to the trained model may accumulate substantially larger degradations over time, whereas decoupling avoids this risk of corrupting the deployed model.

\begin{figure}[h]
    \centering
    \includegraphics[width=1.0\linewidth]{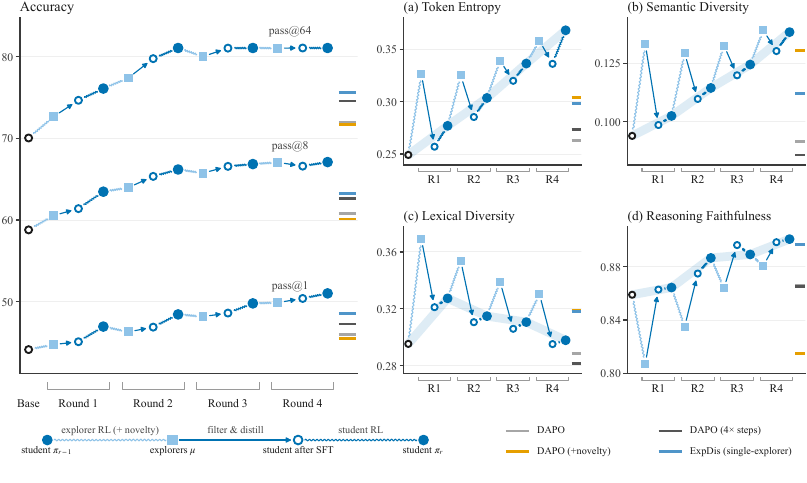}
    \caption{\textbf{In successive ExpDis iterations, the student captures more of the explorer's diversity while avoiding its degradations.} We track the explorers and the student as they evolve during several ExpDis rounds and report the final baseline policies to the right of each panel. \textbf{(a)}~Explorer RL causes a rise in token entropy often at the expense of accuracy, whereas the student's entropy generally rises together with its accuracy across rounds. \textbf{(b)}~The explorer policy semantic diversity stays roughly constant, and the student closes most of the gap to it by the final round. \textbf{(c)}~We find that the student does not retain the lexical diversity of its explorers: it declines over rounds even as the student's accuracy improves. Together with \figref{fig:diversity}, where large $\lambda$ harms the student, this suggests that lexical diversity is not a useful target for exploration. \textbf{(d)}~Reasoning faithfulness, as defined by \citet{rahman2026can}, measures whether a response's reasoning supports its final answer. We find that novelty bonus typically causes faithfulness to decline, though the impact is mild by the final round. Results are averaged over three models (breakdown in \figref{fig:policy-evolution-breakdown}).}
    \label{fig:policy-evolution}
\end{figure}

\paragraph{Policy evolution.} We study how the explorer and student policies evolve across several iterations of ExpDis. \figref{fig:policy-evolution} suggests a simple picture: the explorers repeatedly expand the space of reasoning strategies available to the student, and the student keeps a growing share of that expansion while leaving its costs (reduced pass@$k$ and reasoning faithfulness) behind. Within each round, the novelty bonus pushes the explorer entropy and semantic diversity well above the previous student, and subsequent distillation on the shortest correct trajectories contracts them again. We find that both the student's diversity and accuracy ratchet upward together across subsequent rounds.

Furthermore, explorer policies reach roughly the same semantic diversity in every round, but the gap between them and the student they produce nearly closes by the fourth. One explanation is that each round's explorers start from a stronger student and therefore solve more of the training problems, so more problems contribute a correct trajectory to the distillation set and the student inherits a broader range of the explorer policy's solutions.

\section{Discussion}
\label{sec:discussion}

In standard RLVR, exploration is limited by the need to remain correct. ExpDis removes this constraint by moving exploration into a separate policy. Because the explorer is never deployed, it can be rewarded for novelty, pushed to extremes, or discarded after a single round, with only filtered reasoning traces passing to the student.  Our work is inspired by humans who experiment with bold new ideas when failure is permitted.  We hypothesize that models too need a space to explore, try new ideas, and be allowed to fail.

Our results suggest that decoupling exploration from optimization improves RLVR even at a small budget, and we expect its importance and the value of exploration to grow as RL is scaled. As models improve, they become proficient at employing the solution strategies they have seen before, and the problems worth training on are increasingly those where correct solutions are rare or depart from the solution strategies the model has already mastered. A model optimized only for correctness may avoid sampling such novel strategies. This problem may be especially true beyond mathematical reasoning, for example in long-horizon agentic tasks, theorem proving, and open-ended scientific problems. Decoupling lets explorers be aggressive, and this aggressive exploration might be necessary in order for models to discover completely new ideas.

In our work, we focus our effort primarily on a single novelty bonus and explorers that share the student's architecture. Under ExpDis, explorers could use different novelty signals or sampling strategies to specialize in different parts of the solution distribution. Because an explorer only has to produce useful traces, the best exploration method under ExpDis may look quite different from the best one under standard RLVR. We speculate that exploration methods previously too destabilizing to apply directly to a trained model may become usable in this setting.  ExpDis may unlock new and more aggressive exploration mechanisms.

\section{Acknowledgments}
This project was generously supported by the Google TPU Builders Program, by IBM, and by the NVIDIA Academic Grant Program.
We thank Ang Li, Rohun Agrawal, Kaiwen Zha, Andrea Zanette, and He He for helpful discussions.

\renewcommand{\bibfont}{\small}

\clearpage

\bibliographystyle{plainnat}
\bibliography{references}

\newpage
\appendix

\section{Training \& Hyperparameters}
\label{app:contract}\label{app:training-details}

We report all hyperparameters shared by all methods in \tabref{tab:hparams}. As in the standard DAPO~\citep{yu2025dapo} algorithm, we perform dynamic sampling, overlong filtering, clip-higher,  and do not include a KL penalty term. We use the Dr.~GRPO~\citep{liu2025drgrpo} loss normalization, which divides the token-summed loss by a fixed constant instead of the number of tokens.

\begin{table}[h]
\centering
\caption{\textbf{Hyperparameters.} Shared by all models and all runs unless noted.}
\label{tab:hparams}
\footnotesize
\begin{tabular}{@{}l p{0.64\linewidth}@{}}
\toprule
\multicolumn{2}{l}{\emph{RL (explorer, student, baselines)}} \\
Objective & DAPO clipped objective, $\varepsilon_{\text{low}}{=}0.2$, $\varepsilon_{\text{high}}{=}0.28$, KL $\beta{=}0$ \\
Prompts per step, group size & $B{=}4$, $G{=}16$ \\
Steps & explorer 200, student 100, baselines 300, DAPO ($4\times$ steps) 1{,}200 total \\
Max completion length & 32{,}768 \\
Soft overlong penalty & linear over the final 6{,}554 tokens \\
Optimizer & AdamW ($\beta{=}(0.9, 0.95)$, no weight decay) \\
Learning rate (constant) & explorer and DAPO baselines $5{\times}10^{-6}$; student $1{\times}10^{-6}$ \\
\midrule
\multicolumn{2}{l}{\emph{SFT (student)}} \\
Loss & masked cross-entropy loss (completion-only) \\
Epochs, max examples & 2 epochs, 500 per round \\
Optimizer & AdamW  \\
Learning rate & $5{\times}10^{-6}$ (constant) \\
\midrule
\multicolumn{2}{l}{\emph{Novelty bonus (RND)}} \\
Layers & $\{\lfloor L/4\rfloor, \lfloor L/2\rfloor, \lfloor 3L/4\rfloor\}$ \\
Predictor and target & 3-layer MLP (Linear--ReLU--Linear--ReLU--Linear), width and output 512 \\
Predictor optimizer & Adam, learning rate $10^{-4}$, one step per policy update; predictor and target re-initialized for every explorer and round \\
Features & live policy's hidden states, mean-pooled over completion tokens \\
Bonus & raw (no normalization), credited on verifier-correct completions only \\
\bottomrule
\end{tabular}
\end{table}

\paragraph{RL update.} All RL stages use entirely on-policy steps: vLLM servers reload the trainer's weights after every step and each batch of $4$ prompts $\times$ $16$ samples contributes to the following gradient step.

\begin{figure}[h]
\centering
\includegraphics[width=\linewidth]{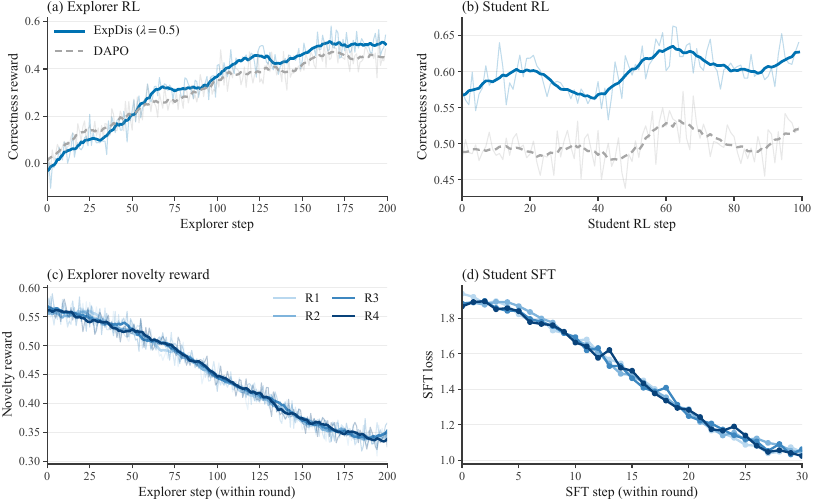}
\caption{Training dynamics of continued training on Qwen3-1.7B with our ExpDis setup ($\lambda{=}0.5$).}
\label{fig:training-grid}
\end{figure}

\subsection{Compute budget}

\begin{table}[h]
\centering
\caption{\textbf{Update allocation per configuration.} MR-ME (multi-round, multi-explorer) combines both axes: $R$ rounds with $K$ explorers per round. Every ExpDis run uses 200 explorer and 100 student updates (19{,}200 selected rollouts); single-model baselines spend all 300 on one model. Parallel explorers run concurrently, so wall-clock time is equal across rows. DAPO ($4\times$ steps) is the only exception.}
\label{tab:budget}
\small
\begin{tabular}{l c l c}
\toprule
Configuration & $R \times K$ & Explorer updates & Student updates per round \\
\midrule
Single-Explorer           & $1 \times 1$ & $200$                         & $100$ \\
Breadth $K{=}2$           & $1 \times 2$ & $100$ each                    & $100$ \\
Breadth $K{=}3$           & $1 \times 3$ & $67/67/66$                    & $100$ \\
Breadth $K{=}5$           & $1 \times 5$ & $40$ each                     & $100$ \\
Breadth $K{=}7$           & $1 \times 7$ & $29/29/29/29/28/28/28$        & $100$ \\
Depth $R{=}4$             & $4 \times 1$ & $50$ per round                & $25$  \\
Depth $R{=}5$             & $5 \times 1$ & $40$ per round                & $20$  \\
MR-ME $R{=}4$, $K{=}3$    & $4 \times 3$ & $17/17/16$ per round          & $25$  \\
MR-ME $R{=}5$, $K{=}3$    & $5 \times 3$ & $14/13/13$ per round          & $20$  \\
MR-ME $R{=}4$, $K{=}5$    & $4 \times 5$ & $10$ per explorer per round   & $25$  \\
\bottomrule
\end{tabular}
\end{table}

\paragraph{Hardware.} Each run uses one TPU v5litepod-64 slice (16 hosts): 4 hosts train the policy with FSDP and 12 serve rollouts with vLLM in bf16 with a 40{,}960-token context.

\begin{table}[h]
\centering
\caption{\textbf{Sampling parameters.} $^\ast$32 for AMC23 and 8 for GSM8K.}
\label{tab:sampling}
\small
\begin{tabular}{l c c}
\toprule
& Training rollouts & Evaluation \\
\midrule
Temperature & 1.0 & 0.6 \\
top-$p$ / top-$k$ / min-$p$ & 0.95 / 20 / -- & 0.95 / 20 / 0 \\
Max completion tokens & 32{,}768 & 32{,}768 \\
Samples per problem & 16 & 64$^\ast$ \\
\bottomrule
\end{tabular}
\end{table}

\subsection{pass@\texorpdfstring{$k$}{k} estimator}
\label{app:estimators}

With $n$ samples per problem, of which $c_i$ are correct on problem $i$, we use the unbiased estimator of \citet{chen2021codex} averaged over $N$ problems:
\begin{equation}
\label{eq:passk}
  \mathrm{pass}@k \;=\; \frac{1}{N}\sum_{i=1}^{N}\left[1 - \frac{\binom{n - c_i}{k}}{\binom{n}{k}}\right], \qquad k \leq n.
\end{equation}

\section{Extended Results}
\label{app:extended-results}

\begin{figure}
    \centering
    \includegraphics[width=0.95\linewidth]{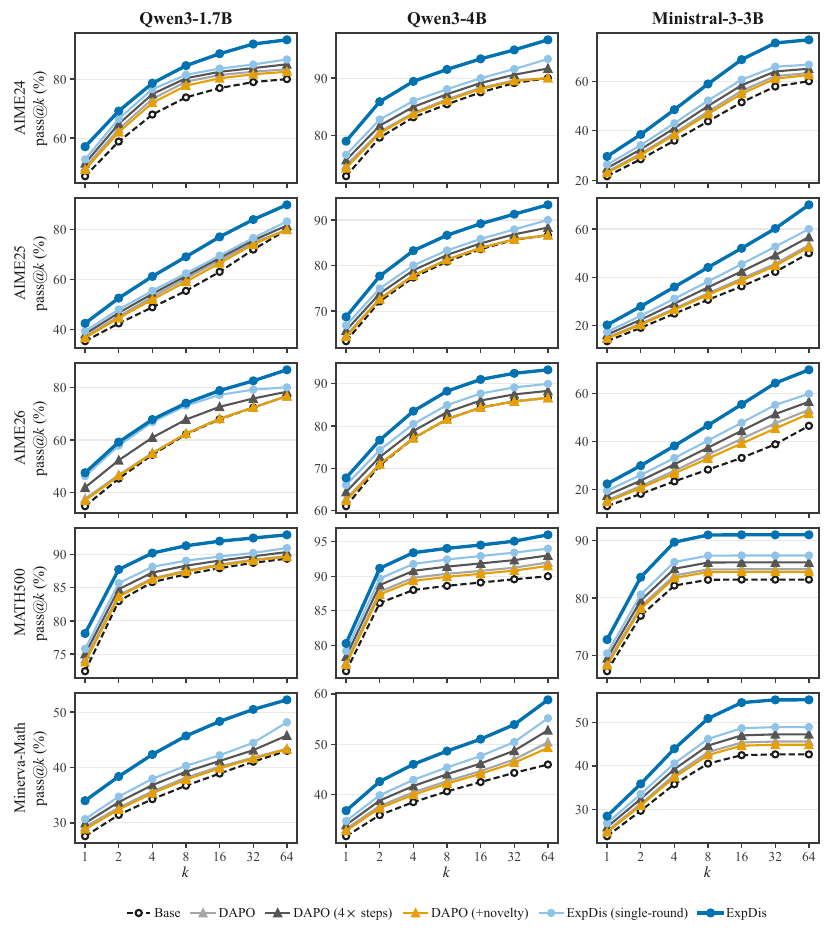}
    \caption{Full pass@$k$ results.}
    \label{fig:full-results}
\end{figure}

We report pass@$k$ for all models and methods in \figref{fig:full-results} and \tabref{tab:passk-mean}.

\label{app:full-tables}

\begin{table}[h]
\centering
\caption{Mean pass@$k$ over the five primary benchmarks.}
  \label{tab:passk-mean}
  \footnotesize
  \setlength{\tabcolsep}{4pt}
  \begin{tabular}{l l rrrrrrr}
  \toprule
  Model & Method & @1 & @2 & @4 & @8 & @16 & @32 & @64 \\
  \midrule
  Qwen3-1.7B & Base & 43.42 & 52.18 & 58.26 & 63.02 & 66.95 & 70.63 & 73.82 \\
  Qwen3-1.7B & DAPO & 45.63 & 54.31 & 60.67 & 65.54 & 69.16 & 72.14 & 74.64 \\
  Qwen3-1.7B & DAPO ($4\times$ steps) & 47.28 & 56.44 & 62.83 & 67.41 & 70.80 & 73.62 & 76.24 \\
  Qwen3-1.7B & DAPO (+novelty) & 45.08 & 53.78 & 60.07 & 64.91 & 68.61 & 71.77 & 74.43 \\
  Qwen3-1.7B & ExpDis (single-round) & 48.93 & 58.55 & 64.98 & 69.27 & 72.43 & 75.10 & 77.83 \\
  Qwen3-1.7B & ExpDis & 51.81 & 61.40 & 68.03 & 72.94 & 76.97 & 80.29 & 83.04 \\
  \midrule
  Qwen3-4B & Base & 61.06 & 68.90 & 72.83 & 75.43 & 77.41 & 78.90 & 79.86 \\
  Qwen3-4B & DAPO & 62.58 & 69.97 & 73.84 & 76.46 & 78.41 & 79.88 & 81.14 \\
  Qwen3-4B & DAPO ($4\times$ steps) & 63.64 & 71.15 & 75.05 & 77.65 & 79.62 & 81.20 & 82.82 \\
  Qwen3-4B & DAPO (+novelty) & 62.20 & 69.70 & 73.59 & 76.20 & 78.16 & 79.64 & 80.82 \\
  Qwen3-4B & ExpDis (single-round) & 64.69 & 72.32 & 76.24 & 78.83 & 80.82 & 82.51 & 84.50 \\
  Qwen3-4B & ExpDis & 66.50 & 74.80 & 79.13 & 81.82 & 83.82 & 85.53 & 87.63 \\
  \midrule
  Ministral-3-3B & Base & 27.95 & 34.54 & 40.52 & 45.34 & 49.36 & 53.00 & 56.50 \\
  Ministral-3-3B & DAPO & 29.63 & 36.63 & 43.21 & 48.79 & 53.45 & 57.19 & 60.12 \\
  Ministral-3-3B & DAPO ($4\times$ steps) & 30.90 & 38.19 & 45.05 & 50.86 & 55.74 & 59.63 & 62.36 \\
  Ministral-3-3B & DAPO (+novelty) & 29.21 & 36.11 & 42.54 & 47.93 & 52.43 & 56.15 & 59.22 \\
  Ministral-3-3B & ExpDis (single-round) & 32.16 & 39.74 & 46.87 & 52.93 & 58.01 & 62.05 & 64.59 \\
  Ministral-3-3B & ExpDis & 34.76 & 43.22 & 51.31 & 58.35 & 64.37 & 69.24 & 72.56 \\
  \bottomrule
  \end{tabular}
  \end{table}

\section{Novelty Bonus Analysis}
\label{app:novelty-bonus}
\noindent
\begin{minipage}[t]{0.48\linewidth}
\vspace{0pt}
\paragraph{Impact on DAPO (+novelty).} In the DAPO (+novelty) baseline, we augment the standard correctness-only reward with a novelty bonus as in
\eqref{eq:scout-reward}. We study how the method performs with various novelty bonus weights $\lambda$ from $0.25$ to
$2$. We perform three training runs over different seeds and find that, with the RND novelty bonus, the mean
mathematical reasoning performance degrades in comparison to DAPO.
\end{minipage}\hfill
\begin{minipage}[t]{0.49\linewidth}
\vspace{-10pt}
\centering
\includegraphics[width=\linewidth]{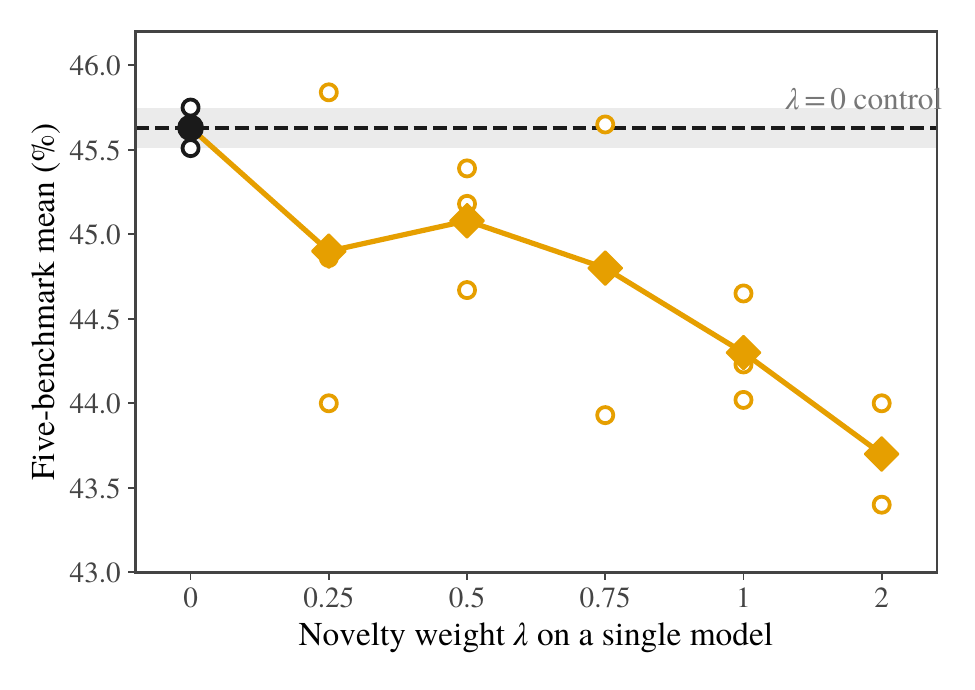}
\captionof{figure}{Adding a novelty bonus directly to the reward in DAPO degrades accuracy on math reasoning.}
\label{fig:direct-novelty-sweep}
\end{minipage}

\paragraph{Impact on student.}
We additionally study the impact that sweeping the novelty bonus weight ($\lambda$) on the explorer policy has on the downstream student policy in a single-round ExpDis setup. Though any amount of novelty bonus hurts the explorer policy, even at high weights (such as $\lambda{=}2$) where pass@$k$ regresses, the downstream student policy does not inherit this degradation.

\begin{figure}[h]
\centering
\includegraphics[width=0.94\linewidth]{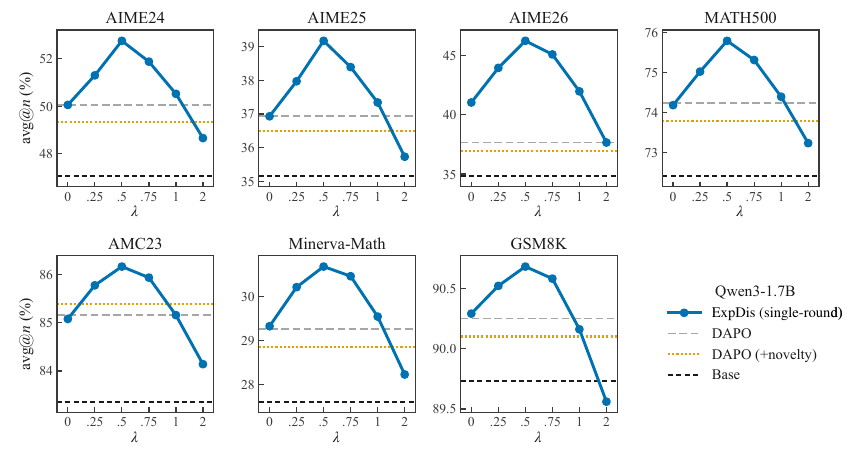}
\caption{On Qwen3-1.7B, we find that a novelty bonus weight of $\lambda$=0.5 on the explorer policy roughly maximizes downstream student policy performance, though several values of $\lambda$ outperform the DAPO baseline.}
\label{fig:lambda-all}
\end{figure}

\begin{figure}[h]
    \centering
    \includegraphics[width=1\linewidth]{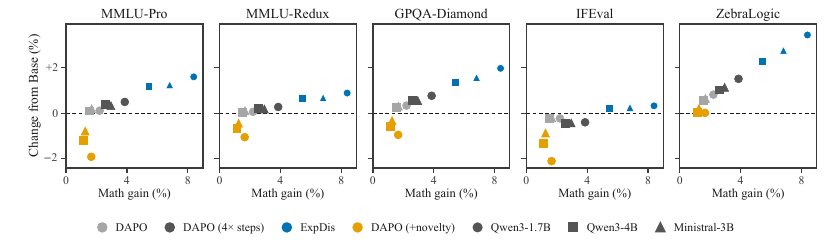}
    \caption{We report the degradation from adding a novelty bonus directly to the DAPO reward.}
    \label{fig:novelty-tax-breakdown}
\end{figure}

\subsection{Novelty metric ($r_\text{novelty}$)}
\label{app:novelty-swap}

To test whether our gains depend on RND, we train two single-round explorers on Qwen3-1.7B with a different novelty term based on the  mean-pooled, L2-normalized final-layer hidden states of each trajectory.

\begin{enumerate}
    \item \emph{kNN novelty} scores each correct completion by its mean cosine distance to its $k{=}16$ nearest neighbors in a replay buffer of the $4{,}096$ most recent correct completions.
    \item \emph{Elliptical novelty}, adapted from the elliptical bonus of \citet{tuyls2025representation}, uses the same features $\phi$ and the bonus $\sqrt{\phi^{\top}\Sigma^{-1}\phi}$, where $\Sigma$ is a ridge-regularized running covariance, $\Sigma = I + \sum_i \phi_i\phi_i^{\top}$ over past correct completions.
\end{enumerate}

We use $\lambda{=}0.5$ for both and report results in \tabref{tab:novelty-swap}.

\begin{table}[H]
\centering
\caption{We ablate the choice of novelty metric used as our novelty bonus.}
\label{tab:novelty-swap}
\small
\setlength{\tabcolsep}{4.5pt}
\begin{tabular}{l l rrrrr r}
\toprule
Method & $r_{\text{novelty}}$ & AIME24 & AIME25 & AIME26 & MATH500 & Minerva & Mean \\
\midrule
DAPO                & --         & 50.05 & 36.93 & 37.66 & 74.24 & 29.27 & 45.63 \\
DAPO ($4\times$ steps) & --      & 51.41 & 38.05 & 41.96 & 75.02 & 29.98 & 47.28 \\
\midrule
ExpDis (ours)       & RND        & 52.76 & 39.17 & 46.25 & 75.80 & 30.69 & 48.93 \\
ExpDis              & kNN        & 50.83 & 37.50 & 43.75 & 75.00 & 29.75 & 47.37 \\
ExpDis              & Elliptical & 51.56 & 38.23 & 44.79 & 75.30 & 30.40 & 48.06 \\
\bottomrule
\end{tabular}
\end{table}

\begin{figure}[h]
\centering
\includegraphics[width=\linewidth]{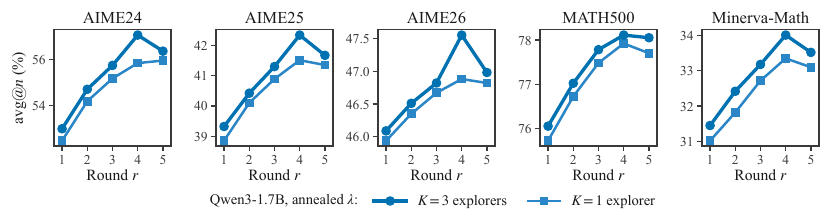}
\caption{Impact of increasing rounds given a fixed number of explorers per round.}
\label{fig:rounds-perbench}
\end{figure}

\section{Diversity Analysis}
\label{app:diversity}

We define the diversity metrics used in the paper.

\paragraph{InterDistinct-4.} Given $n$ generations $\{g_1,\dots,g_n\}$ for a single problem, let $\mathrm{4grams}(g_i)$ be the multiset of $4$-grams in $g_i$ and $G_4 = \bigcup_i \mathrm{4grams}(g_i)$ the set of distinct $4$-grams in the pool. Then
\[
  \text{InterDistinct-4} \;=\; \frac{|G_4|}{\sum_i |\mathrm{4grams}(g_i)|},
\]
the ratio of distinct $4$-grams to total $4$-grams in the pool. We average across problems.

\paragraph{AnswerDistinct@$n$.} For each problem, count the number of distinct extracted final answers across the $n$ generations, then average across problems.

\paragraph{Token entropy.} For each model we sample 8 rollouts for each of the 30 AIME24 problems with the evaluation sampler and record the mean per-token entropy of the sampled tokens.

\paragraph{Semantic diversity.} We follow \citet{li2025darling}, who finetune Qwen3-Embedding-4B as a classifier of whether two mathematical reasoning traces use the same approach. We use their checkpoint to produce pairwise similarity scores between all generations for a single prompt. We then cluster the generations and report the resulting number of clusters $C$ normalized by the number of generations.

\begin{table}[H]
\centering
\caption{We report diversity metrics for Qwen3-1.7B models on AIME24 generations.}
\label{tab:released-diversity}
\footnotesize
\setlength{\tabcolsep}{3pt}
\begin{tabular}{l r r r r r}
\toprule
Model & Entropy & Semantic & Ans.\ entropy & Distinct ans. & InterDistinct-4 \\
\midrule
Base                     & $0.237$ & 0.100 & 1.42 & 8.94 & 0.311 \\
DAPO                     & $0.251$ & 0.096 & 1.37 & 8.50 & 0.308 \\
DAPO ($4\times$ steps)   & $0.265$ & 0.091 & 1.35 & 8.46 & 0.302 \\
ExpDis (single-explorer) & $0.306$ & 0.120 & 1.41 & 9.93 & 0.357 \\
ExpDis                   & $0.391$ & 0.143 & 1.30 & 9.44 & 0.339 \\
\bottomrule
\end{tabular}
\end{table}

\begin{table}[h]
\centering
\caption{We report InterDistinct-4 (Inter4) and AnswerDistinct@$n$ (Ans@$n$) for the explorer and the student on AIME24 across explorer weights $\lambda$.}
\label{tab:diversity-rebased}
\scriptsize
\setlength{\tabcolsep}{3pt}
\begin{tabular}{l l rr rr}
\toprule
& & \multicolumn{2}{c}{Explorer} & \multicolumn{2}{c}{Student} \\
\cmidrule(lr){3-4} \cmidrule(l){5-6}
Model & Method & Inter4 & Ans@$n$ & Inter4 & Ans@$n$ \\
\midrule
Qwen3-1.7B & Base                                  & --    & --    & 0.311 & 8.94  \\
Qwen3-1.7B & DAPO                                  & --    & --    & 0.308 & 8.50  \\
Qwen3-1.7B & DAPO (+novelty, $\lambda{=}0.5$)      & --    & --    & 0.342 & 9.39  \\
Qwen3-1.7B & ExpDis ($\lambda{=}0.25$)             & 0.403 & 11.60 & 0.346 & 9.63  \\
Qwen3-1.7B & ExpDis ($\lambda{=}0.5$)              & 0.422 & 12.61 & 0.357 & 9.93  \\
Qwen3-1.7B & ExpDis ($\lambda{=}0.75$)             & 0.478 & 14.44 & 0.363 & 10.02 \\
Qwen3-1.7B & ExpDis ($\lambda{=}1.0$)              & 0.534 & 16.28 & 0.335 & 9.30  \\
Qwen3-1.7B & ExpDis ($\lambda{=}2.0$)              & 0.550 & 23.61 & 0.283 & 7.87  \\
\midrule
Qwen3-4B   & Base                                  & --    & --    & 0.290 & 4.02  \\
Qwen3-4B   & DAPO                                  & --    & --    & 0.287 & 3.82  \\
Qwen3-4B   & DAPO (+novelty, $\lambda{=}0.5$)      & --    & --    & 0.319 & 4.22  \\
Qwen3-4B   & ExpDis ($\lambda{=}0.5$)              & 0.394 & 5.67  & 0.333 & 4.47  \\
\bottomrule
\end{tabular}
\end{table}

\begin{figure}
    \centering
    \includegraphics[width=1\linewidth]{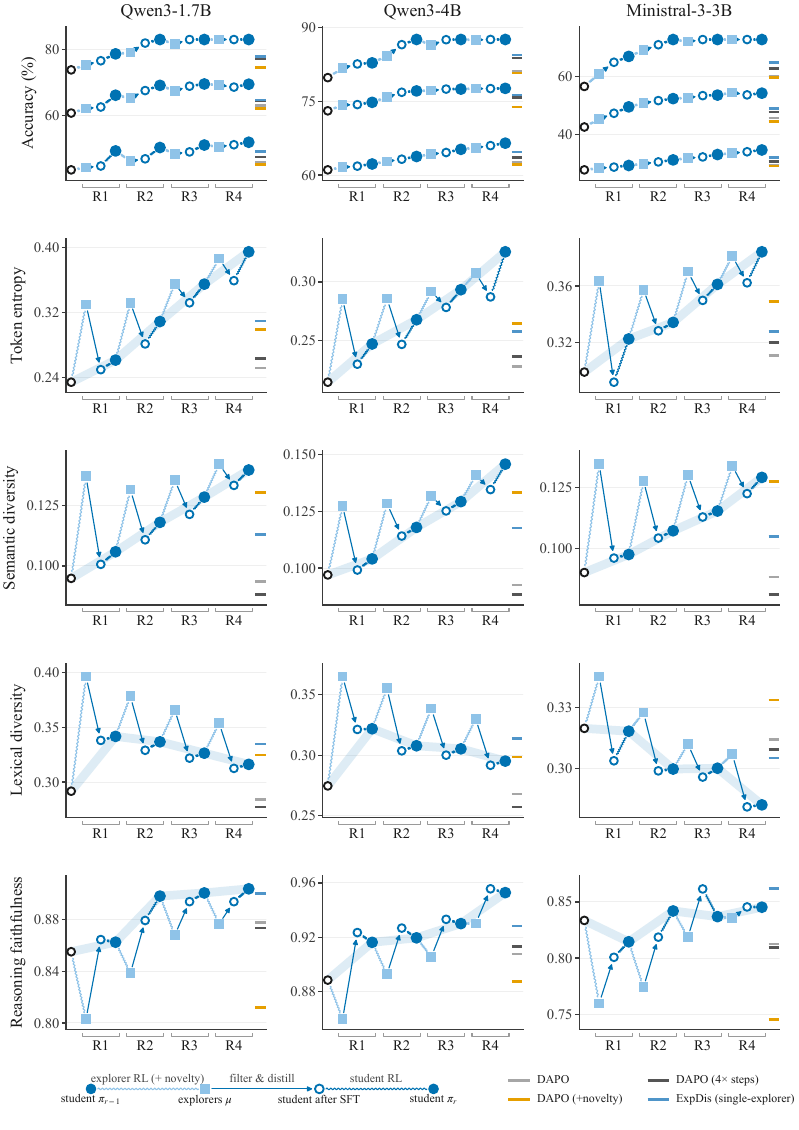}
    \caption{We study the evolution of student and teacher policies during several stages of ExpDis.}
    \label{fig:policy-evolution-breakdown}
\end{figure}

\section{Ablations}
\label{app:ablations}

\paragraph{Quality filtering.} With three explorers, keeping every correct trajectory from the pooled explorers
(NaivePool) gives a mean of $48.95$\%, no better than a single explorer ($48.93$\%). Applying the quality filter increases performance to $49.47$\%.

\begin{figure}[h]
\centering
\begin{minipage}[c]{0.48\linewidth}
\centering
\captionof{table}{Ablating student distillation, student RL, or both.}
\label{tab:filter}
\small
\begin{tabular}{l r}
\toprule
Configuration & AIME24 \\
\midrule
DAPO & 50.05 \\
Unfiltered SFT + RL & 49.79 \\
Filtered SFT, no RL & 49.84 \\
Filtered SFT + RL (ExpDis) & 52.76 \\
\bottomrule
\end{tabular}
\end{minipage}\hfill
\begin{minipage}[c]{0.46\linewidth}
\centering
\includegraphics[width=\linewidth]{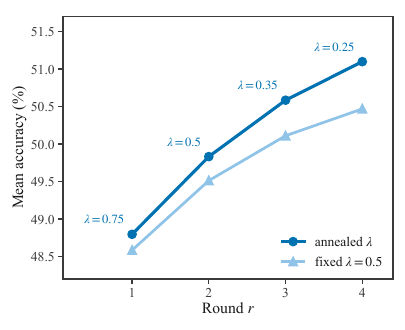}
\caption{Fixed versus annealed $\lambda$ across four rounds (Qwen3-1.7B, single-explorer). The annealed schedule uses $\lambda{=}0.75$, $0.5$, $0.35$, and $0.25$.}
\label{fig:lambda-ablation}
\end{minipage}
\end{figure}

\paragraph{Student distillation vs student RL.} On Qwen3-1.7B, distilling unfiltered explorer traces and then running RL reaches
$49.79$\% on AIME24, and distilling filtered traces without RL reaches $49.84$\%. Both variants underperform DAPO (at $50.05$\%). Only filtering
followed by RL reaches $52.76$\% (\tabref{tab:filter}).

\paragraph{Annealing $\lambda$.} We find that annealing $\lambda$ during multiple rounds ($0.75, 0.5, 0.35, 0.25$) outperforms a fixed $\lambda{=}0.5$ at every round (see \figref{fig:lambda-ablation}).

\begin{figure}[tbp]
\centering
\includegraphics[width=0.92\linewidth]{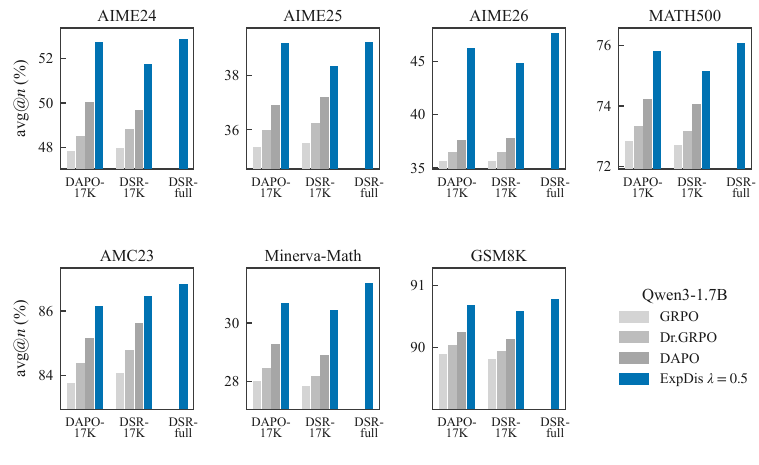}
\caption{We ablate the choice of training dataset, sampling a fixed prompt budget from each and training with a single-explorer ExpDis approach. We find a consistent gain over several RLVR baselines.}
\label{fig:dataset}
\end{figure}

\paragraph{Training data.} Every run samples the same number of prompts (about $1{,}200$), so these runs differ only in
the pool they sample from (\figref{fig:dataset}, \tabref{tab:dataset}). Sampling from DAPO-Math-17K or a 17K subset of
DeepScaleR gives similar results, with neither better on every benchmark. Sampling ExpDis prompts from the full 40K
DeepScaleR pool is better on every benchmark. All main results use DAPO-Math-17K.

\begin{table}[h]
\centering
\caption{We ablate the choice of training dataset. We see roughly the same performance across all datasets.}
\label{tab:dataset}
\footnotesize
\setlength{\tabcolsep}{3.5pt}
\begin{tabular}{l l rrrrrrr}
\toprule
Method & Dataset & AIME24 & AIME25 & AIME26 & MATH500 & AMC23 & Minerva & GSM8K \\
\midrule
GRPO    & DAPO-Math-17K  & 47.81 & 35.36 & 35.68 & 72.84 & 83.75 & 28.02 & 89.89 \\
GRPO    & DeepScaleR-17K & 47.96 & 35.52 & 35.68 & 72.71 & 84.09 & 27.86 & 89.81 \\
Dr.~GRPO & DAPO-Math-17K  & 48.49 & 35.99 & 36.46 & 73.34 & 84.38 & 28.45 & 90.04 \\
Dr.~GRPO & DeepScaleR-17K & 48.83 & 36.26 & 36.51 & 73.18 & 84.78 & 28.20 & 89.95 \\
DAPO    & DAPO-Math-17K  & 50.05 & 36.93 & 37.66 & 74.24 & 85.16 & 29.27 & 90.25 \\
DAPO    & DeepScaleR-17K & 49.68 & 37.21 & 37.76 & 74.08 & 85.62 & 28.90 & 90.14 \\
\midrule
ExpDis $\lambda{=}0.5$ & DAPO-Math-17K  & 52.76 & 39.17 & 46.25 & 75.80 & 86.17 & 30.69 & 90.68 \\
ExpDis $\lambda{=}0.5$ & DeepScaleR-17K & 51.77 & 38.33 & 44.84 & 75.16 & 86.48 & 30.44 & 90.59 \\
ExpDis $\lambda{=}0.5$ & DeepScaleR-full & 52.89 & 39.24 & 47.66 & 76.08 & 86.84 & 31.37 & 90.78 \\
\bottomrule
\end{tabular}
\end{table}

\end{document}